\documentclass[letterpaper, 10 pt, conference]{ieeeconf}
\IEEEoverridecommandlockouts

\usepackage[colorlinks]{hyperref}
\hypersetup{
    colorlinks,
    linkcolor=black,
    citecolor=black,
    filecolor=magenta,
    urlcolor=cyan,
}

\usepackage{resizegather}

\usepackage{cite}
\usepackage{amsmath,amssymb,amsfonts,mathrsfs}
\usepackage{graphicx}
\usepackage{textcomp}
\usepackage{xcolor}
\usepackage{tcolorbox}
\usepackage{textcomp}
\usepackage{tablefootnote}
\usepackage{threeparttable}
\usepackage{caption, subcaption}
\usepackage{bbm}
\usepackage{dsfont}
\usepackage{tikz}
\usepackage{graphicx}
\usepackage{tkz-euclide}
\usepackage{algorithm}
\usepackage{algpseudocode}
\usetikzlibrary{positioning}
\usetikzlibrary{shapes}
\usetikzlibrary{shapes.misc}
\usetikzlibrary{shapes.geometric}
\usetikzlibrary{plotmarks}
\usetikzlibrary{intersections}
\usetikzlibrary{calc}
\usetikzlibrary{fit}
\usetikzlibrary{patterns,tikzmark}
\usetikzlibrary{matrix,decorations.pathreplacing,calc}

\tikzset{cross/.style={cross out, draw, 
         minimum size=2*(#1-\pgflinewidth), 
         inner sep=0pt, outer sep=0pt}}

\newcommand{\state}[0]{x}

\newcommand{\norm}[1]{\left\lVert#1\right\rVert}

\newcommand{\vertiii}[1]{{\left\vert\kern-0.25ex\left\vert\kern-0.25ex\left\vert #1 
    \right\vert\kern-0.25ex\right\vert\kern-0.25ex\right\vert}}

\newtheorem{theorem}{Theorem}

\newtheorem{lemma}{Lemma}

\newtheorem{problem}{Problem}

\newtheorem{remark}{Remark}

\newtheorem{definition}{Definition}

\makeatletter
\renewcommand{\fps@figure}{htp}
\renewcommand{\fps@table}{htp}
\makeatother

\def\BibTeX{{\rm B\kern-.05em{\sc i\kern-.025em b}\kern-.08em
    T\kern-.1667em\lower.7ex\hbox{E}\kern-.125emX}}
    
\begin{document}

\title{Learning a Formally Verified Control Barrier Function \\ in Stochastic Environment}

\author{Manan Tayal$^{1}$, Hongchao Zhang$^{2}$, Pushpak Jagtap$^{1}$, Andrew Clark$^{2}$, Shishir Kolathaya$^{1}$ 
\thanks{
}
\thanks{$^{1}$Cyber-Physical Systems, Indian Institute of Science (IISc), Bengaluru.
{\tt\scriptsize \{manantayal, pushpak, shishirk\}@iisc.ac.in}
.
}%
\thanks{$^{2}$ Electrical \& Systems Engineering, Washington University in St. Louis.
{\tt\scriptsize \{hongchao, andrewclark\}@wustl.edu}
.
}
}

\maketitle
\begin{abstract}
Safety is a fundamental requirement of control systems. Control Barrier Functions (CBFs) are proposed to ensure the safety of the control system by constructing safety filters or synthesizing control inputs. However, the safety guarantee and performance of safe controllers rely on the construction of valid CBFs. Inspired by universal approximatability, CBFs are represented by neural networks, known as neural CBFs (NCBFs).
This paper presents an algorithm for synthesizing formally verified continuous-time neural Control Barrier Functions in stochastic environments in a single step. The proposed training process ensures efficacy across the entire state space with only a finite number of data points by constructing a sample-based learning framework for Stochastic Neural CBFs (SNCBFs). Our methodology eliminates the need for post hoc verification by enforcing Lipschitz bounds on the neural network, its Jacobian, and Hessian terms. We demonstrate the effectiveness of our approach through case studies on the inverted pendulum system and obstacle avoidance in autonomous driving, showcasing larger safe regions compared to baseline methods.
\end{abstract}
\begin{keywords}
     control barrier function, data driven controls, stochastic environments, neural networks
\end{keywords}

\section{Introduction}
\label{section: Introduction}
In the rapidly evolving landscape of control theory, ensuring safety in real-world applications has emerged as a paramount concern. With the pervasive integration of automated systems such as self-driving cars, the ability to guarantee their safety becomes increasingly critical. Current methods, including traditional optimal control approaches and Hamilton-Jacobi reachability analysis, have been instrumental in addressing safety requirements by casting them as constraints or generating safe controls offline. However, the scalability limitations of these methods, coupled with the challenge of swift reactions in complex environments, underscore the need for alternative approaches to safety assurance.
With the rise in popularity of learning-based methods in control synthesis, learning-based methods have also been used to synthesize controllers for safety-critical tasks. However, most learning-based methods require a significant amount of unsafe interactions to learn a safe controller, which might be costly or impossible to obtain.

One promising avenue in addressing safety concerns is through the utilization of Control Barrier Functions (CBFs)\cite{ames2014control}. CBFs serve as a practical method for synthesizing safe control for control affine systems\cite{Ames_2017}\cite{ames2019control} as well as stochastic control systems \cite{jagtap2020formal,clark2021verification}. By formulating controllers through Quadratic Programs (QPs), solvable at high frequencies with modern optimization solvers, CBFs have found applications in various safety-critical tasks, including adaptive cruise control~\cite{ames2014control}, aerial maneuvers \cite{7525253,tayal2023control} and legged locomotion~\cite{ames2019control,nguyen2015safety}. In these applications, the performance and safety guarantees depend on the CBF that is used. While control synthesis techniques such as sum-of-squares~\cite{1470374,TOPCU20082669} have been used to construct polynomial control barrier functions, they have been limited to systems with low-dimensional state spaces.




In recent years, the emergence of neural network-based barrier function (NCBF) synthesis has garnered considerable attention due to the universal approximation property of neural networks. A variety of methods have been proposed for training NCBFs, including learning from expert demonstrations \cite{9303785}, SMT-based techniques \cite{zhao2020synthesizing,abate2021fossil,abate2020formal,10.1007/978-3-030-72016-2_20}, mixed-integer programs \cite{zhao2022verifying}, and nonlinear programs \cite{NEURIPS2023_120ed726}. Loss functions for training NCBFs were proposed in \cite{dawson2022safe,dawson2023safe,liu2023safe,9993334,zhang2024fault}, while  \cite{so2023train} synthesizes CBFs by learning a value function of a nominal policy and demonstrating that the maximum-over-time cost serves as a CBF. In these existing works, the trained CBFs must be verified as a postprocessing step and recomputed if verification fails, resulting in a potentially time-consuming trial-and-error process. 

More recently, a data-driven approach for training NCBFs was proposed in \cite{anand2023formally}. The proposed training process also bounds the Lipschitz constant of the trained NCBF. By exploiting these Lipschitz bounds, the safety of the continuous state space can be verified using only a discrete set of sample points. This approach, however, only considers discrete-time deterministic systems. In a continuous-time setting, the safety guarantees also depend on the Lie derivative of the NCBF, and hence Lipschitz bounds on the NCBF alone are insufficient to guarantee safety. Moreover, the abovementioned works on NCBFs do not consider the presence of stochastic system noise, making formally verifiable synthesis of stochastic NCBFs an open research problem.

In this paper, we propose an algorithm to synthesize a formally verified continuous-time neural CBF in stochastic environments in a single step. 
We construct a sample-based learning framework to train Stochastic Neural CBF (SNCBF) and prove the trained SNCBF ensuring efficacy across the entire state space with only a finite number of data points. To summarize, this paper makes the following contributions.
\begin{itemize}
    \item We propose a training framework to synthesize provably correct Control Barrier Functions (CBFs) parameterized as neural networks for continuous-time, stochastic systems, eliminating any need for post hoc verification.
    \item Our methodology establishes completeness guarantees by deriving a validity condition, ensuring efficacy across the entire state space with only a finite number of data points. We train the network by enforcing Lipschitz bounds on the neural network, its Jacobian and (trace of) Hessian terms.
    \item We evaluate our approach using two case studies, namely, the inverted pendulum system and the obstacle avoidance of an autonomous driving system. We show that our training framework successfully constructs an SNCBF to differentiate safe and unsafe regions. We demonstrate our proposed method ensures a larger safe region compared with the baseline method in \cite{zhang2024fault}. 
\end{itemize}

The rest of this paper is organized as follows. Preliminaries explaining the concept of control barrier functions (CBFs) and safety filter designs are introduced in Section \ref{section: Background}. Problem formulation is discussed in Section \ref{section: Problem Formulation}. The method proposed to synthesize the CBFs using neural networks, the construction of loss functions and the algorithm to train it, are discussed in Section \ref{section: Method}. The simulation results will be discussed in Section \ref{section: Simulation}. Finally, we present our conclusion in Section \ref{section: Conclusions}.


\section{Preliminaries}
\label{section: Background}
In this section, we will formally introduce Control Barrier Functions (CBFs) and their importance for real-time safety-critical control.
\subsection{Notations}
\label{subsec:notations}
We first present the definitions of class-$\mathcal{K}$ and extended class-$\mathcal{K}$ functions. A continuous function $\kappa : [0, d) \rightarrow [0, \infty)$ for some $d > 0$ is said to belong to \emph{class-$\mathcal{K}$} if it is strictly increasing and $\kappa(0) = 0$. Here, $d$ is allowed to be $\infty$. The same function can extended to the interval $\kappa: (-b,d)\to (-\infty, \infty)$ with $b>0$ (which is also allowed to be $\infty$), in this case we call it the \emph{extended class-$\mathcal{K}$ function}. Given a square matrix $A$, the trace representing the sum of its diagonal elements is denoted by $\mathsf{tr}(A)$, and the determinant is denoted by $\operatorname{det}(A)$. Given a function $\phi(x)$, we represent the derivative and double derivative of $\phi(x)$ with respect to the input $x$ as $\phi'(x)$ and $\phi''(x)$, respectively. If $m = [m_1, \dots, m_n]^T\in\mathbb{R}^n$ is a $n \times 1$ matrix, then $\mathsf{diag}(m) = \begin{bmatrix} 
    m_1 & \dots & 0\\
    \vdots & \ddots & \vdots\\
    0 & \dots  & m_{n}
    \end{bmatrix}$.

\subsection{System Description}
\label{subsec:model}

We consider a continuous time stochastic control system with state $x(t) \in \mathcal{X}\subseteq\mathbb{R}^{n}$ and input $u(t) \in \mathbb{U} \subseteq \mathbb{R}^{m}$ at time $t\geq0$. The state dynamics is described by the stochastic differential equation as:
\begin{align}
    \label{eq:state-sde}
    dx(t) &= \big(f(x(t))+g(x(t))u(t)\big) \ dt + \sigma \ dW(t),
\end{align}
where functions $f: \mathbb{R}^{n} \rightarrow \mathbb{R}^{n}$ and $g: \mathbb{R}^{n} \rightarrow \mathbb{R}^{n \times m}$ are locally Lipschitz, $\sigma \in \mathbb{R}^{n \times n}$, $W$ is an $n$-dimensional Brownian motion. 

Consider a set $\mathcal{C}$ defined as the \textit{super-level set} of a continuously differentiable function $h:\mathcal{X}\subseteq \mathbb{R}^n \rightarrow \mathbb{R}$ yielding,
\begin{align}
\label{eq:setc1}
	\mathcal{C}                        & = \{ \state \in \mathcal{X} \subset \mathbb{R}^n : h(\state) \geq 0\} \\
\label{eq:setc2}
	\mathcal{X}-\mathcal{C} & = \{ \state \in \mathcal{X} \subset \mathbb{R}^n : h(\state) < 0\}.
\end{align}
We further let the interior and boundary of $\mathcal{C}$ be $\text{Int}\left(\mathcal{C}\right) = \{ \state \in \mathcal{X} \subset \mathbb{R}^n : h(\state) > 0\}$ and $\partial\mathcal{C} = \{ \state \in \mathcal{X} \subset \mathbb{R}^n : h(\state) = 0\}$, respectively. 

We define a Lipschitz continuous control policy $\mu: \mathbb{R}^n \rightarrow \mathbb{R}^m$ to be a mapping from the sequence of states $x(t)$ to a control input $u(t)$ at each time $t$. The safety of a controlled system is defined as follows. 

\begin{definition}[Safety]
    \label{def:positive-invariance}
    A set $\mathcal{C} \subseteq \mathcal{X} \subseteq \mathbb{R}^{n}$ is positive invariant under dynamics \eqref{eq:state-sde} and control policy $\mu$ if $x(0)\in \mathcal{C}$ and $u(t) = \mu(x(t)) \ \forall t \geq 0$ imply that $x(t) \in \mathcal{C}$ for all $t \geq 0$. 
    If $\mathcal{C}$ is positive invariant, then the system satisfies the safety constraint with respect to $\mathcal{C}$.
\end{definition}


\subsection{Control Barrier Functions (CBFs)}

The control policy needs to guarantee the robot satisfies a safety constraint, which is specified as the positive invariance of a given safety region $\mathcal{C}$. The Control Barrier Function (CBF) are widely used to synthesize a control policy with positive invariance guarantees. We next present the definition of CBFs as discussed in \cite{ames2019control} for nonstochastic systems.

\begin{definition}[Control barrier function (CBF)]{
\label{definition: CBF definition}
Given a control-affine system $\dot x=f(x)+g(x)u$, the set $\mathcal{C}$ defined by \eqref{eq:setc1}, with $\frac{\partial h}{\partial \state}(\state) \neq 0$ for all $\state \in \partial \mathcal{C}$, the function $h$ is called the control barrier function (CBF) defined on the set $\mathcal{X}$, if there exists an extended \textit{class}-$\mathcal{K}$ function $\kappa$ such that for all $\state \in \mathcal{X}$:
\begin{equation}
\begin{aligned}
    \underset{ u \in \mathbb{U}}{\text{sup}}\! \left[\underbrace{\mathcal{L}_{f} h(\state) + \mathcal{L}_g h(\state)u} \iffalse+ \frac{\partial h}{\partial t}\fi_{\dot{h}\left(\state, u\right)} \! + \kappa\left(h(\state)\right)\right] \! \geq \! 0,
\end{aligned}
\end{equation}
where $\mathcal{L}_{f} h(\state) = \frac{\partial h}{\partial \state}f(\state)$ and $\mathcal{L}_{g} h(\state)= \frac{\partial h}{\partial \state}g(\state)$ are the Lie derivatives.}
\end{definition}

By \cite{Ames_2017} and \cite{ames2019control}, we have that any Lipschitz continuous control law $\mu(\state)$ satisfying the inequality: $\dot{h} + \kappa( h )\geq 0$ ensures safety of $\mathcal{C}$ if $x(0)\in \mathcal{C}$, and asymptotic convergence to $\mathcal{C}$ if $x(0)$ is outside of $\mathcal{C}$. 
 
We next present the safety lemma to ensure the safety of continuous-time stochastic systems. 

\begin{lemma}[\cite{clark2021verification}]
    The set $\mathcal{C} \subset \mathbb{R}^n$ be a set defined on the super-level set of a continuously differentiable function $h : \mathcal{X} \subset \mathbb{R}^n \rightarrow \mathbb{R}$. The function $h$ defined on the set $\mathcal{X}$ is a CBF for stochastic system in \eqref{eq:state-sde}, if there exists an extended \textit{class}-$\mathcal{K}$ function $\kappa$ such that for all $\state \in \mathcal{X}$:
    \begin{equation}
    \label{eq: lie_derivative}
    \begin{aligned}
        \underset{ u \in \mathbb{U}}{\text{sup}}\! \left[\mathcal{L}_{f} h(\state) + \mathcal{L}_g h(\state)u +
        \frac{1}{2}\mathsf{tr}\left(\sigma^\intercal \frac{\partial^2 h(\state)}{\partial x^2} \sigma\right)+ \kappa\left(h(\state)\right)\right] \! \geq \! 0.
    \end{aligned}
    \end{equation}
\end{lemma}


\subsection{Controller Synthesis for Real-time Safety}
\label{subsection: safe_controller}
Having described the CBF and its associated formal results, we now discuss its Quadratic Programming (QP) formulation. 
CBFs are typically regarded as \textit{safety filters} which take the desired input (reference controller input) $u_{ref}(\state,t)$ and modify this input in a minimal way: 

\begin{equation}
\begin{aligned}
\label{eqn: CBF QP}
u^{*}(x,t) &= \min_{u \in \mathbb{U} \subseteq \mathbb{R}^m} \norm{u - u_{ref}(x,t)}^2\\
\quad & \textrm{s.t. } \mathcal{L}_f h(x) + \mathcal{L}_g h(x)u + 
    \frac{1}{2}\mathsf{tr}\left(\sigma^\intercal \frac{\partial^2 h(\state)}{\partial x^2} \sigma\right) \\
    & + \kappa \left(h(x)\right) \geq 0.
\end{aligned}
\end{equation}
This is called the Control Barrier Function based Quadratic Program (CBF-QP).

\section{Problem Formulation}
\label{section: Problem Formulation}
In this section, we begin by formally defining the problem of synthesis of CBF in a stochastic environment and highlighting the challenges associated with its direct solution. Subsequently, we propose a reformulation of the problem and derive the conditions, that provide formal guarantees on the correctness of the solution over the entire state set despite using finitely many data samples.

To leverage the universal approximation property of neural networks, we represent the Stochastic Control Barrier Function with a feed-forward neural network. Denoted as $\tilde{h}(x \mid\theta)$, where $\theta$ signifies the trainable parameters of the neural network, this representation is termed the Stochastic Neural CBF ($\mathrm{SNCBF}$). Due to the absence of an $\mathrm{SNCBF}$, we lack access to the safe set $\mathcal{C}$. Hence, we start with initial safe and unsafe sets $\mathcal{X}_s \subseteq \mathcal{C}$ and $\mathcal{X}_u \subseteq \mathcal{X}-\mathcal{C}$, respectively, such that any trajectory starting in $\mathcal{X}_s$ never enters $\mathcal{X}_u$. We next formulate the problem.

\begin{problem}
\label{prob:origin}
     Given a continuous-time stochastic control system defined as \eqref{eq:state-sde}, state set $\mathcal{X}$, initial safe and unsafe sets $\mathcal{X}_{s}$ and $\mathcal{X}_{u}$, respectively, the objective is to devise an algorithm to synthesize $\mathrm{SNCBF}$ $\tilde{h}(x \mid\theta)$ using a Lipschitz continuous controller $u$ such that 
    \begin{align}
        & \tilde{h}(x \mid \theta) \geq 0, \forall x\in \mathcal{X}_s, \notag\\
        & \tilde{h}(x \mid\theta) < 0,  \forall x\in \mathcal{X}_u, \notag\\
        & \frac{\partial \tilde{h}\left(x \mid \theta\right)}{\partial x} (f(x)+g(x) u(x)) +\frac{1}{2}\mathsf{tr}\left(\sigma^\intercal \frac{\partial^2 \tilde{h}\left(x \mid \theta\right)}{\partial x^2} \sigma\right) \notag\\
        & +  \kappa\left(\tilde{h}\left(x \mid \theta\right)\right) \geq 0,  \forall x\in \mathcal{X}.\label{eq: problem1}
    \end{align}
\end{problem}

In order to enforce conditions \eqref{eq: problem1} in Problem \ref{prob:origin}, we first cast our problem as the following robust optimization problem (ROP):
\begin{equation}\label{eq: rcp}
    \mathrm{ROP}: \begin{cases}\underset{\psi}{\min } & \psi \\ \text { s.t. } & \max \left(q_{k}(x)\right) \leq \psi, k \in\{1,2,3\} \\ & \forall x \in \mathcal{X}, \psi \in \mathbb{R},\end{cases}
\end{equation}

where
\begin{equation} \label{eq:q_conditions}
    \begin{aligned}
        q_{1}(x)=& \left(-\tilde{h}(x \mid \theta)\right) \mathds{1}_{\mathcal{X}_{s}}, \\
         q_{2}(x)=& \left(\tilde{h}(x \mid \theta)+\delta \right) \mathds{1}_{\mathcal{X}_{u}}, \\
         q_{3}(x)=&  -\frac{\partial \tilde{h}\left(x \mid \theta\right)}{\partial x} (f(x)+g(x) u(x))  \\
        & -\frac{1}{2}\mathsf{tr}\left(\sigma^\intercal \frac{\partial^2 \tilde{h}\left(x \mid \theta\right)}{\partial x^2} \sigma\right) -  \kappa\left(\tilde{h}\left(x \mid \theta\right)\right),
    \end{aligned}
\end{equation}
where $\delta$ is a small positive value to ensure the strict inequality. If the optimal solution of the ROP ($\psi_{\mathrm{ROP}}^{*}$) $\leq 0$, then the conditions \eqref{eq: problem1} are satisfied and the $\mathrm{SNCBF}$ is valid. 

However, the proposed ROP in \eqref{eq: rcp} has infinitely many constraints since the state of the system in a continuous set. 
This motivates us to employ data-driven approaches and the scenario optimization program of ROP. 
Given $\bar{\epsilon}$, suppose we sample  $N$ data points: $x_{i} \in \mathcal{X}, i \in\{1, \ldots, N\}$, such that $\left\|x-x_{i}\right\| \leq \bar{\epsilon}$. Instead of solving the ROP in \eqref{eq: rcp}, we employ the following scenario optimization problem (SOP):
\begin{equation}\label{eq: scp}
    \mathrm{SOP}: \begin{cases}\underset{\psi}{\min } & \psi \\ \text { s.t. } & q_{1}\left(x_{i}\right) \leq \psi, \forall x_{i} \in \mathcal{S}, \\ & q_{2}\left(x_{i}\right) \leq \psi, \forall x_{i} \in \mathcal{U}, \\ & q_{3}\left(x_{i}\right) \leq \psi, \forall x_{i} \in \mathcal{D}, \\ & \psi \in \mathbb{R}, i \in\{1, \ldots, N\},\end{cases}
\end{equation}
where $q_{k}(x), k \in\{1,2,3\}$ are defined as in \eqref{eq:q_conditions}. The data sets $\mathcal{S}, \mathcal{U}$ and $\mathcal{D}$ corresponding to points sampled from the initial safe set $\mathcal{X}_s \subseteq \mathcal{C}$, initial unsafe set $\mathcal{X}_u \subseteq \mathcal{X}-\mathcal{C}$, and state set $\mathcal{X}$, respectively.


Given the finite number of data samples $x_{i}$, and considering SOP as a linear program in relation to the decision variable $\psi$, it becomes feasible to find a solution for the SOP. Let us denote the optimal solution of the SOP as $\psi^{*}$.
We now derive conditions under which the $\mathrm{SNCBF}$ $\tilde{h}(x \mid \theta)$
satisfies conditions \eqref{eq: problem1}. The following theorem shows that solving \eqref{eq: rcp} can be achieved by finding a solution to \eqref{eq: scp}. 

\begin{theorem}\label{thm: }{
Consider a continuous time stochastic control system \eqref{def:positive-invariance}, and initial safe and unsafe sets $\mathcal{X}_{s} \subseteq \mathcal{X}$ and $\mathcal{X}_{u} \subseteq \mathcal{X}$, respectively. Let  $\tilde{h}(x \mid \theta)$ be the neural network-based CBF with trainable parameters $\theta$. For the SOP \eqref{eq: scp} constructed by utilizing $N$ samples, let $\psi^{*}$ be the optimal value, with the assumption that functions $q_{k}(x), k \in\{1,2,3\}$ in equation \eqref{eq:q_conditions} are Lipschitz continuous. Then $\tilde{h}(x \mid \theta)$ is a valid $\mathrm{SCBF}$, i.e., it satisfies conditions \eqref{eq: problem1}, if the following condition holds:}
\begin{equation} \label{eq: completeness_condition}
    L_{\max} \bar{\epsilon}+\psi^{*} \leq 0,
\end{equation}
\textit{where $L_{\max}$ is maximum of the Lipschitz constants of $q_{k}(x), k \in\{1,2,3\}$ in \eqref{eq:q_conditions}.}
\end{theorem}

\begin{proof}
For any $x$ and any $k\in\{1,2,3\}$, we know that:
\begin{equation*} 
    \begin{aligned}
    q_k(x) & =  q_k(x) - q_k(x_i) + q_k(x_i)\\
    & \leq L_{k}\left\|x - x_i \right\| + \psi^*\\
    & \leq L_{k}\bar{\epsilon} + \psi^* \leq L_{\max}\bar{\epsilon} + \psi^* \leq 0.
    \end{aligned}
\end{equation*}
 Hence, if $q_{k}(x), k \in\{1,2,3\}$ satisfies condition \eqref{eq: completeness_condition}, then the $\tilde{h}(x \mid \theta)$ is a valid $\mathrm{CBF}$, satisfying conditions \eqref{eq: problem1}.
\end{proof}

Hence, the original Problem \ref{prob:origin} can be efficiently addressed by solving the problem reformulated as follows. 


\begin{problem}
\label{prob:reform}
     {Given a continuous-time stochastic control system defined as \eqref{eq:state-sde} and the data sets $\mathcal{S}, \mathcal{U}$ and $\mathcal{D}$, the objective is to devise an algorithm to synthesize $\mathrm{SNCBF}$ $\tilde{h}(x \mid\theta)$ and consequently, $\mathrm{SNCBF-QP}$ based controller $u$, such that they satisfy the conditions required in SOP \eqref{eq: scp} (functions $q_{k}(x) \leq \psi^*, k \in\{1,2,3\}$ in equation \eqref{eq:q_conditions}) and $\psi^*$ satisfies condition \eqref{eq: completeness_condition}.}
\end{problem}

\section{Method}
\label{section: Method}
In this section, we propose an algorithmic approach to solve the problem formulated in Section \ref{section: Problem Formulation}. The structure of this section is as follows. We first present the method to synthesize SNCBF to solve Problem \ref{prob:reform} and then demonstrate the training process. 



\subsection{Synthesis of SNCBF}

Following the problem formulation in Section \ref{section: Problem Formulation}, we now describe the construction of suitable loss functions for the training  $\mathrm{SNCBF}$ $\tilde{h}(x \mid \theta)$ such that its minimization leads to the solution of Problem \ref{prob:reform}. As described in the previous section, $\mathrm{SNCBF}$ is a feed-forward neural network, with trainable weight parameters $\theta$. To address the aforementioned problem, it is imperative to compute $\frac{\partial \tilde{h}\left(x \mid \theta\right)}{\partial x}$ and $\mathsf{tr}(\frac{\partial^2 \tilde{h}\left(x \mid \theta\right))}{\partial x^2}$. This necessitates the selection of a neural network with a smooth activation function, thereby facilitating the derivation of smooth Jacobian and Hessian values upon differentiation. Interested readers are directed to \cite{lutter2019deep}, which analytically computes the Jacobian and Hessian using the chain rule.


Now, let us consider the following loss functions satisfying the conditions required by SOP \eqref{eq: scp} over the training data sets $\mathcal{S}, \mathcal{U}, \mathcal{D}$ as follows:
\begin{equation}
    \begin{aligned}
        & \mathcal{L}_{1}(\theta)=\frac{1}{N} \sum_{x_{i} \in \mathcal{S}} \max \left(0,q_1(x_{i}) - \psi \right), \\
        & \mathcal{L}_{2}(\theta)=\frac{1}{N} \sum_{x_{i} \in \mathcal{U}} \max \left(0, q_2(x_{i}) - \psi\right), \\
        & \mathcal{L}_{3}(\theta)= \frac{1}{N} \sum_{x_{i} \in \mathcal{D}} \max \left(0, q_3(x_{i}) - \psi \right)
    \end{aligned}
\end{equation}
with $\mathcal{L}_{1}$ representing the loss for safe states, $\mathcal{L}_{2}$ representing the loss for unsafe states and $\mathcal{L}_{3}$ representing the loss for lie derivative conditions over the entire state set, respectively. 
Using the terms defined above, the loss function is given by
\begin{equation}\label{eq: total_loss}
    \mathcal{L}_{\theta}(\theta)=\mathcal{L}_{1}+\lambda_{1} \mathcal{L}_{2}+\lambda_{2} \mathcal{L}_{3},
\end{equation}
with $\lambda_{1}, \lambda_{2} \in \mathbb{R}_{+}$weighting the importance of the individual loss terms.

To ensure the fulfillment of the assumptions in Theorem \ref{thm: }, it is imperative to verify the Lipschitz boundedness of the functions $q_{k}(x)$, where $k \in\{1,2,3\}$. This necessitates the Lipschitz boundedness of $\tilde{h}(x \mid \theta)$, $\frac{\partial \tilde{h}\left(x_{i} \mid \theta\right)}{\partial x}$, and $\mathsf{tr}(\sigma^\intercal \frac{\partial^2 \tilde{h}\left(x_{i} \mid \theta\right)}{\partial x^2} \sigma)$, with corresponding Lipschitz bounds denoted as $L_{h}$, $L_{dh}$, and $L_{d2h}$, respectively.
To train neural networks with Lipschitz bounds, we have the following lemma:
\begin{lemma}[\cite{pauli2021training}]\label{lemma: lmi_const}
{Suppose $f_{\theta}$ is a $l$-layered feed-forward neural network with $\theta$ as a trainable parameter, then a certificate for $L$-Lipschitz continuity of the neural network is given by the semi-definite constraint $M(\theta, \Lambda) :=$} \\
\begin{equation*}
\begin{aligned}
    \begin{bmatrix} 
    A\\
    B
    \end{bmatrix}^T \begin{bmatrix} 
    2 \alpha\beta\Lambda & -(\alpha + \beta)\Lambda\\
    -(\alpha + \beta)\Lambda & 2\Lambda
    \end{bmatrix}
    \begin{bmatrix} 
    A\\
    B
    \end{bmatrix} \\
    + \begin{bmatrix} 
    L^2 \textbf{\textit{I}} & 0  & 0  & 0\\
    0 & 0 & 0 & 0\\
    0 & 0 & 0 & -\theta_l^T\\
    0 & 0 & -\theta_l &  \textbf{\textit{I}}
    \end{bmatrix} \succeq 0,
\end{aligned}
\end{equation*}
where 
\begin{equation*}
    A = \begin{bmatrix} 
    \theta_0 & \dots & 0  & 0\\
    \vdots & \ddots & \vdots & \vdots\\
    0 & \dots  & \theta_{l-1} & 0
    \end{bmatrix}, 
    B = \begin{bmatrix} 
    0 & I
    \end{bmatrix},
\end{equation*}
{$(\theta_0, \dots \theta_l) $ are the weights of the neural network, $\Lambda \in \mathbb{D}_{+}^{n}$, $i=1,\dots,l$ and $\alpha$ and $\beta$ are the minimum and maximum slopes of the activation functions, respectively. }
\end{lemma}

\begin{remark}
{
For a single-layer case, the matrix $M$ reduces to:
\begin{equation*}
    M(\theta, \Lambda) =
    \begin{bmatrix} 
    L^2 \textbf{\textit{I}} + 2 \alpha\beta\theta_0^T\Lambda\theta_0 & -(\alpha + \beta)\theta_0^T\Lambda  & 0\\
    -(\alpha + \beta)\Lambda\theta_0 & 2\Lambda & -\theta_1\\
    0 & -\theta_1 &  \textbf{\textit{I}}
    \end{bmatrix} \succeq 0.
\end{equation*}
}
\end{remark}

The lemma above addresses the certification of the L-Lipschitz bound for a neural network. However, our scenario necessitates ensuring not only the Lipschitz boundedness of the SNCBF $\tilde h$, but also of $\frac{\partial \tilde{h}\left(x_{i} \mid \theta\right)}{\partial x}$ and $\sigma^\intercal \frac{\partial^2 \tilde{h}\left(x_{i} \mid \theta\right)}{\partial x^2} \sigma$. Therefore, we must explore the relationship between the network weights and the semi-definite matrix $M$ to guarantee the boundedness of the aforementioned terms. To address this issue, we introduce the following theorem.

\begin{theorem} \label{thm: effective_weights}
    {
    Consider a $1$-layered feedforward neural network $f_{\theta}$, with output dimension $1 \times 1$, where $\theta$ represents the trainable weight parameters. Let $y$ denote the output of the neural network, $x$ denote the input of the neural network, $\theta_i$, $i \in {0,1}$, denote the weight parameters of each layer with $\theta = \{\theta_0, \theta_1\}$ and $\phi$ denote the activation. The certificate for  L-Lipschitz continuity of the derivative of the neural network $(\frac{\partial y}{\partial x})$ is given by $M_{\hat\phi}(\hat\theta, \Lambda)\succeq0$, where $\hat\phi= \phi'$ and  $\hat\theta = (\theta_0, \hat\theta_1)$, $\hat\theta_1$ is defined as:
    \begin{equation}
        \hat\theta_1 = \theta_{0}^T \mathsf{diag}(\theta_{1}).
    \end{equation}

    Similarly, the certificate for the L-Lipschitz continuity of $\mathsf{tr}(\sigma^T\frac{\partial^2 y_l}{\partial x^2}\sigma)$ is expressed as $M_{\bar\phi}(\bar\theta, \Lambda)\succeq0$, where $\bar\phi= \phi''$ and $\bar\theta = (\theta_0,\bar\theta_1)$ and $\bar\theta_1$ is defined as:
    \begin{equation}
        \bar\theta_1 = \begin{bmatrix} \sum_{j=0}^{r}\sigma_j^2\theta_1^{j1}\theta_0^{1j} & \dots & \sum_{j=0}^{r}\sigma_j^2\theta_1^{jn}\theta_0^{pj}.
    \end{bmatrix}
    \end{equation}
    }
\end{theorem}

\begin{proof}
   The proof can be found in the Appendix. 
\end{proof}

We address a constrained optimization problem aiming to minimize loss $\mathcal{L}\left(f_{\theta}\right)$ subject to $\boldsymbol{M}_{j}(\theta, \Lambda) \succeq 0$ for $j=0, \ldots, p$, where $f_{\theta}$ is a given neural network. By employing a log-det barrier function, we convert this into an unconstrained optimization problem:
\begin{equation*}
\min _{\theta, \Lambda} \mathcal{L}\left(f_{\theta}\right) + \mathcal{L}_{M}(\theta, \Lambda),
\end{equation*}
where $\mathcal{L}_{M}(\theta, \Lambda) = -\sum_{j=0}^{q} \rho_{j} \log \operatorname{det}\left(\boldsymbol{M}_{j}(\theta, \Lambda)\right)$ and $\rho_{j}>0$ are barrier parameters. Ensuring that the loss function $\mathcal{L}_{M}(\theta, \Lambda) \leq 0$, guarantees that the linear matrix inequalities $\boldsymbol{M}_{j}(\theta, \Lambda) \succeq 0, j=1, \ldots, p$ hold true. 
Let us consider the loss functions characterizing the satisfaction of Lipschitz bound as
\begin{equation}\label{eq: lmi_loss}
    \begin{aligned}
     \mathcal{L}_{M}&(\theta, \Lambda, \hat\Lambda, \bar\Lambda)=-c_{l_{1}} \log \operatorname{det}(M_1(\theta, \Lambda)) \\ &-c_{l_{2}} \log \operatorname{det}(M_2(\hat\theta, \hat\Lambda))
      -c_{l_{3}} \log \operatorname{det}(M_3(\bar\theta, \bar\Lambda)),
    \end{aligned}
\end{equation}
where $c_{l_{1}}, c_{l_{2}}, c_{l_{3}}$ are positive weight coefficients for the sub-loss functions, $M_1, M_2, M_3$ are the semi-definite matrices corresponding to the Lipschitz bounds $L_{h}, L_{dh}, L_{d2h}$ respectively, $\Lambda, \hat\Lambda, \bar\Lambda$ are trainable parameters and $\theta, \hat\theta, \bar\theta$ are the weights mentioned in Theorem \ref{thm: effective_weights}.

Finally, let us consider the following loss function to satisfy validity condition \eqref{eq: completeness_condition}:
\begin{equation}\label{eq: completeness_loss}
    \begin{aligned}
    \mathcal{L}_{v}(\psi)=\max \left(0,L_{\max } \bar{\epsilon}+\psi\right),
    \end{aligned}
\end{equation}
where $L_{\max}$ is maximum of the Lipschitz constants of $q_{k}(x), k \in\{1,2,3\}$ in \eqref{eq:q_conditions}, or $L_{\max}=\max \left(L_{h}, L_{h}+L_{dh}L_{x} + L_{d2h}\right)$.

\subsection{Training with Safety Guarantee}

\begin{algorithm}
\caption{Learning Formally Verified Stochastic Neural Control Barrier Functions}\label{alg:sncbf}
\begin{algorithmic}
\Require Data Sets: $\mathcal{S}, \mathcal{U}, \mathcal{D}$, Dynamics: $f, g, \sigma$, Lipschitz Bounds: $L_h, L_{dh}, L_x, L_{d2h}$
\State Initialise($\theta, \psi, \Lambda, \Lambda', \Lambda''$)
\State $x_i \gets sample(\mathcal{S}, \mathcal{U},\mathcal{D})$
\State $L_{\max} \gets L_h, L_{dh}, L_x, L_{d2h}$
\While{$\mathcal{L}_{\theta} > 0$ or $\mathcal{L}_M \not\leq 0$ or $\mathcal{L}_v > 0 $ }
    \State $\tilde h \gets \theta$
    \State $u_i \gets \mathrm{SNCBF-QP}(\tilde h, f, g, \sigma, x_i, \psi)$ \Comment{From eq. \eqref{eq: sncbf-qp}}
    \State $\mathcal{L}_{\theta} \gets (\tilde h, f, g, \sigma, x_i, u_i, \psi)$ \Comment{From eq. \eqref{eq: total_loss}}
    \State $\theta \gets Learn(\mathcal{L}_{\theta}, \theta)$
    \State $\mathcal{L}_{M} \gets (\theta, \Lambda, \Lambda', \Lambda'')$ \Comment{From eq. \eqref{eq: lmi_loss}}
    \State $\theta, \Lambda, \Lambda', \Lambda'' \gets Learn(\mathcal{L}_{M})$
    \State $\mathcal{L}_{v} \gets (L_{\max}, \psi)$ \Comment{From eq. \eqref{eq: completeness_loss}}
    \State $\psi \gets Learn(\mathcal{L}_{v})$
\EndWhile
\end{algorithmic}
\end{algorithm}

The overall training procedure is as follows. We first fix all the hyper-parameters required for the training, including $\bar{\epsilon}, \mathcal{L}_{h}, \mathcal{L}_{dh}, \mathcal{L}_{d2h}, \lambda_{1}, \lambda_{2}, c_{l_{1}}, c_{l_{2}}, c_{l_{3}}$, and the maximum number of epochs considered.
At each time step, the controller intakes the current state $x$ and reference control input $u_{\text{ref}}(x)$. The basic idea is to construct safe control input by filtering out unsafe actions based on the condition derived from learned $\mathrm{SNCBF}$. Specifically, the controller minimizes the modification on the control input $u$ compared to $u_{\text{ref}}$ such that the $\mathrm{SNCBF}$ safety condition holds by solving $\mathrm{SNCBF-QP}$ defined as follows. 
    \begin{align}
        &\min _{u \in \mathcal{U}}  \left\|u-u_{\text {ref }}(x)\right\| \notag\\
        & \ \ \textrm{s.t.} \frac{\partial \tilde{h}\left(x_{i} \mid \theta\right)}{\partial x} (\mathbf{f}(x_{i})+\mathbf{g}(x_{i}) u)  + \frac{1}{2}\mathsf{tr}\left(\sigma^\intercal \frac{\partial^2 \tilde{h}\left(x_{i} \mid \theta\right)}{\partial x^2} \sigma\right) \notag\\
        &\ \ \ \ \  \geq-\kappa(\tilde{h}(x_{i})).\label{eq: sncbf-qp}
    \end{align}
Finally, the control action $u$ can be used in computing loss functions in the next iteration. The training data sets are randomly shuffled into several batches, and the loss is calculated for each batch at a time. Then, using a stochastic gradient descent algorithm, e.g., ADAM,  the trainable parameters $\theta, \Lambda, \hat\Lambda,\bar\Lambda$ and $\psi$ are updated. This procedure is repeated until the network converges or if a predefined maximum episode number is reached. For our proposed approach, all of the learning is done offline in a simulation environment. After the learning process converges, the SNCBF-QP based controller $u$ can then be deployed to the intended system. The overall algorithm is summarised in Algorithm \ref{alg:sncbf}

We next present the following theorem that provides formal guarantees of safety for the continuous time stochastic control system by utilizing the trained SNCBF $\tilde{h}(x \mid \theta)$.

\begin{theorem}
    \textit{Consider a continuous time stochastic control system \eqref{def:positive-invariance}, and safe and unsafe sets $X_{s} \subseteq X$ and $X_{u} \subseteq X$, respectively. Let  $\tilde{h}(x \mid \theta)$ be the trained $\mathrm{SNCBF}$, such that $\mathcal{L}_{\theta} = 0$, $\mathcal{L}_M \leq 0$ and $\mathcal{L}_v = 0$. Then, the control action $u$ obtained by solving the $\mathrm{SNCBF-QP}$ \eqref{eq: sncbf-qp}, makes the system safe.}
\end{theorem} 
\begin{proof}
   When $\mathcal{L}_M \leq 0$, it indicates that the Lipschitz bound on the SNCBF $\tilde h$, $\frac{\partial \tilde{h}\left(x{i} \mid \theta\right)}{\partial x}$, and $\sigma^\intercal \frac{\partial^2 \tilde{h}\left(x_{i} \mid \theta\right)}{\partial x^2} \sigma$ is satisfied. Consequently, we can compute $L_{\max}$, which is necessary to validate the conditions outlined in \eqref{eq: completeness_condition}.Moreover, when $\mathcal{L}_v = 0$, it signifies the fulfillment of the validity conditions \eqref{eq: completeness_condition}, leading to the determination of an optimal $\phi^*$. Similarly, $\mathcal{L}_{\theta} = 0$ denotes the satisfaction of the conditions stipulated in SOP \eqref{eq: scp} (functions $q_{k}(x) \leq \psi^*, k \in{1,2,3}$ in equation \eqref{eq:q_conditions}). Consequently, all the conditions outlined in \eqref{eq: problem1} are met, resulting in the formal verification of the SNCBF and ensuring the safety of the system through the controller $u$ derived by solving the SNCBF-QP.
\end{proof}

\section{Simulations}
\label{section: Simulation}
In this section, we assess the efficacy of our proposed methodology through two distinct case studies: the inverted pendulum model and the obstacle avoidance of an autonomous mobile robot \cite{barry2012safety}. Both case studies are conducted on a computing platform equipped with an Intel i7-12700H CPU, 32GB RAM, and NVIDIA GeForce RTX 3090 GPU.  For the class $\mathcal{K}$ function in the CBF inequality  \eqref{eq: lie_derivative}, we chose $\kappa(h) = \gamma h$, where $\gamma=1$. 
\begin{figure*}[htp]
\begin{subfigure}{.3\textwidth}
    \centering
    \includegraphics[width=\textwidth]{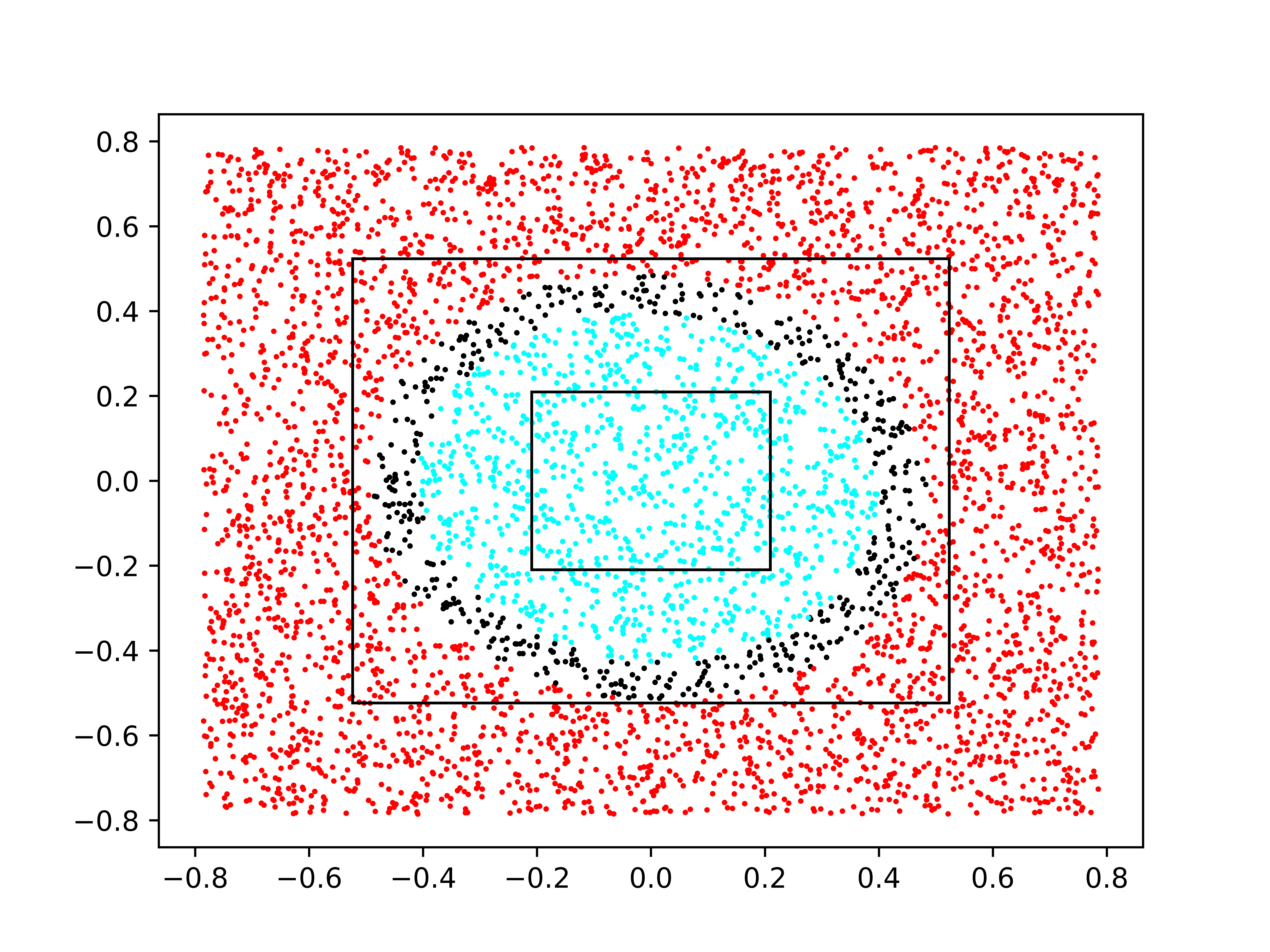}
    \subcaption{}
    \label{fig:2d_sample}
\end{subfigure}%
\begin{subfigure}{.36\textwidth}
    \centering
    \includegraphics[width=\textwidth]{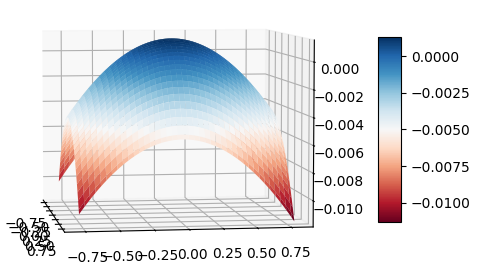}
    \subcaption{}
    \label{fig:3d_sncbf}
\end{subfigure}%
\begin{subfigure}{.26\textwidth}
    \centering
    \includegraphics[width=\textwidth]{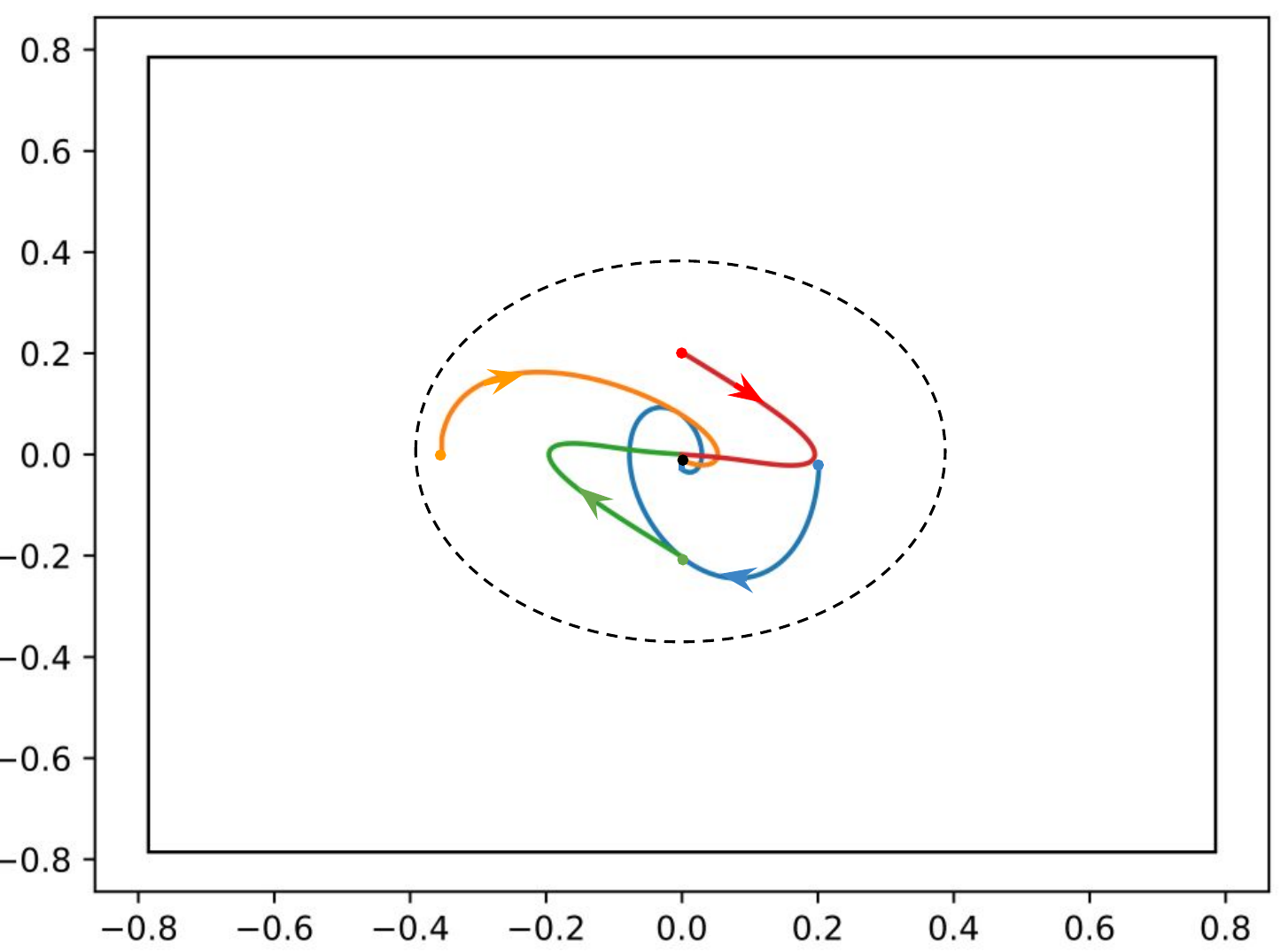}
    \subcaption{}
    \label{fig:ip_traj} 
\end{subfigure}
\caption{This figure presents the experimental results on the inverted pendulum system.
Fig. \ref{fig:2d_sample} visualizes of $\tilde{h}\left(x \mid \theta\right)$ over $X$. Blue and red regions denote the safe region ($\tilde{h} \geq -\psi^*$) and the unsafe region ($\tilde{h} < \psi^*$), respectively. The initial safety region boundary and unsafe region boundary is denoted by black boxes. We observe that the boundary of trained SNCBF (black dots) successfully separate the unsafe and safe region. 
Fig. \ref{fig:3d_sncbf} shows the 3D plot of $\tilde{h}\left(x \mid \theta\right)$ over $X$. We observe that the safe region is greater than zero while unsafe region has negative function value. 
Fig. \ref{fig:ip_traj} presents trajectories initiating inside the safe set following SNCBF-QP, following different reference controllers
}
\end{figure*}

\subsection{Inverted Pendulum}
We consider a continuous time stochastic inverted pendulum dynamics given as follows:
\begin{equation}
    d\left[\begin{array}{l}
    \theta \\
    \dot{\theta}
    \end{array}\right]=\left(\left[\begin{array}{c}
    \dot{\theta} \\
    \frac{g}{l}\sin(\theta)
    \end{array}\right]+\left[\begin{array}{c}
    0 \\
    \frac{1}{m l^{2}} 
    \end{array}\right] u\right) dt + \sigma \ dW_{t},
\end{equation}
where $\theta \in \mathbb{R}$ denotes the angle, $\dot{\theta} \in \mathbb{R}$ denotes the angular velocity, $u \in \mathbb{R}$ denotes the controller, $m$ denotes the mass and $l$ denotes the length of the pendulum. We let the mass $m= 1$kg, length $l=10$m and the disturbance $\sigma = \mathsf{diag}(0.1, 0.1)$. 
The inverted pendulum operate in a state space given as $\mathcal{X}=\left[-\frac{\pi}{4}, \frac{\pi}{4} \right]^{2}$. 
The system is required to stay in a limited safe stable region.  
The initial safe region is given as $\mathcal{X}_{s}=\left[-\frac{\pi}{15}, \frac{\pi}{15}\right]^{2}$ and the unsafe region is given as $\mathcal{X}_{u}=\mathcal{X} \backslash \left[-\frac{\pi}{6}, \frac{\pi}{6}\right]^{2}$. 

We train the SNCBF, assuming knowledge of the model. The SNCBF $\tilde{h}$ consists of one hidden layer of $20$ neurons, with Softplus activation function ($log(1+exp(x)$). 




We set the training hyper-parameters to $\bar{\epsilon}=0.00016$, $L_{h} = 0.01$, $L_{dh} = 0.4$ and $L_{d2h} = 2$ yielding $L_{\max} = 2.4$. We perform the training to simultaneously minimize the loss functions $\mathcal{L}_{\theta}, \mathcal{L}_{M}$, and $\mathcal{L}_{v}$. The training algorithm then converges to obtain the SNCBF $\tilde{h}\left(x \mid \theta\right)$ with $\psi^{*}=-0.00042$. Thus, using Theorem \ref{thm: }, we can verify that the SNCBF thus obtained is valid, thus ensuring safety. 

Visualizations of the trained SNCBF are presented in both 2D with sample points (Fig. \ref{fig:2d_sample}) and in 3D function value heat map (Fig. \ref{fig:3d_sncbf}). 
These visualizations demonstrate the successful separation of the initial safety region boundary from the unsafe region boundary. We validate our SNCBF-QP based safe controller on an inverted pendulum model on PyBullet. Fig \ref{fig:ip_traj} shows the trajectories initiating inside the safe set (with different reference controllers) and never leaving the safe set, thus validating our approach.

\subsection{Obstacle Avoidance}

We consider an autonomous mobile robot navigating on a road following the dynamics \cite{dubins1957curves} given below
\begin{equation}
    d\left[\begin{array}{l}
    x_1 \\
    x_2 \\
    \psi
    \end{array}\right]=\left(\left[\begin{array}{c}
    v\cos\psi \\
    v\sin\psi \\
    0
    \end{array}\right]+\left[\begin{array}{c}
    0 \\
    0 \\
    1 
    \end{array}\right] u\right) dt + \sigma \ dW_{t}
\end{equation}
where $[x_1, x_2, \psi]^T\in\mathcal{X}\subseteq\mathbb{R}^3$ is the state consisting of the location $(x_1,x_2)$ of the robot and its orientation $\psi$, with $v$ representing the robot's speed and $u$ controlling its orientation. We set the speed $v = 1$ and the disturbance $\sigma = \mathsf{diag}(0.1, 0.1, 0.1)$.

\begin{figure}[t]
    \centering
\vspace{-0.6em}
\includegraphics[width=0.9\linewidth]{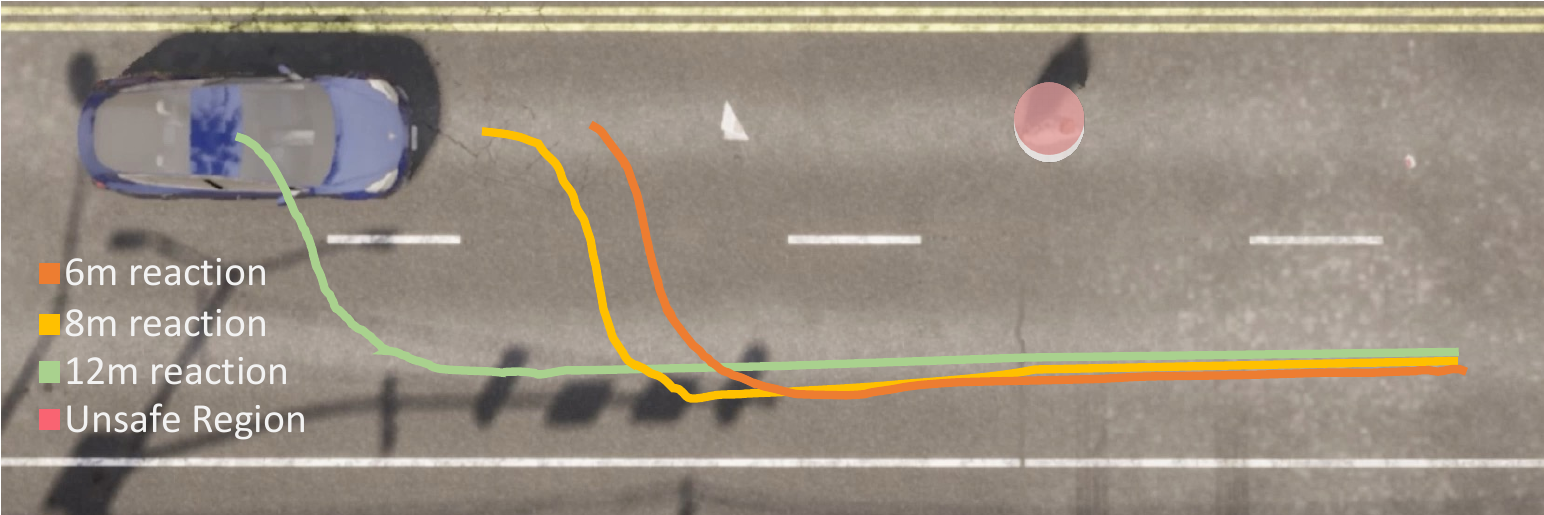}
    \caption{Proposed safe control comparison among different reaction distance. We let the vehicle to adjust its orientation to maneuver in its lane. We show three trajectories to demonstrate our proposed SNCBF-based controller under different initial state, namely, $6$, $8$ and $12$ meter away from the pedestrian, respectively. Three trajectories of the vehicle under control shows our proposed method succeeds in maneuvering the vehicle to avoid the pedestrian. }
    \label{fig:carla_obs_avoid}
\end{figure}

The mobile robot is required to stay on the road while avoiding pedestrians sharing the field of activities. The unicycle operates in a state space given as $\mathcal{X}= [-2,2]^3$. The initial safe region is given as $\mathcal{X}_{s}=\mathcal{X} \backslash [-1.5, 1.5]^{2} \times [-2,2]$ and the unsafe region is given as $\mathcal{X}_{u}=[-0.2, 0.2]^{2} \times [-2,2]$. 

We train the SNCBF, assuming knowledge of the model with the same architecture as before (one hidden layer of $20$ neurons, with Softplus activation function ($log(1+exp(x)$)). 
The training hyper-parameters are set to $\bar{\epsilon}=0.01$, $L_{h} = 1$, $L_{dh} = 1$ and $L_{d2h} = 2$ resulting in $L_{\max} = 4$. We perform the training to simultaneously minimize the loss functions $\mathcal{L}_{\theta}, \mathcal{L}_{M}$, and $\mathcal{L}_{v}$. The training algorithm then converges to obtain the SNCBF $\tilde{h}\left(x \mid \theta\right)$ along with $\psi^{*}=-0.04002$.

We validate our safe control in the CARLA simulation environment, in which the vehicle maneuvers its orientation to conduct obstacle avoidance. If the control policy guides the vehicle to another lane with no obstacle ahead, then it switches to a baseline autonomous driving algorithm. The trajectory of the vehicle is shown in Fig. \ref{fig:carla_obs_avoid}. We show the trajectory of the proposed SNCBF-based safe control in different reaction distances, namely, $6$, $8$, and $12$ meters, respectively. All three trajectories of the vehicle under control show our proposed method succeeds in maneuvering the vehicle to avoid the pedestrian. 

\begin{figure}[t]
    \centering
\vspace{-0.6em}
\includegraphics[width=0.9\linewidth]{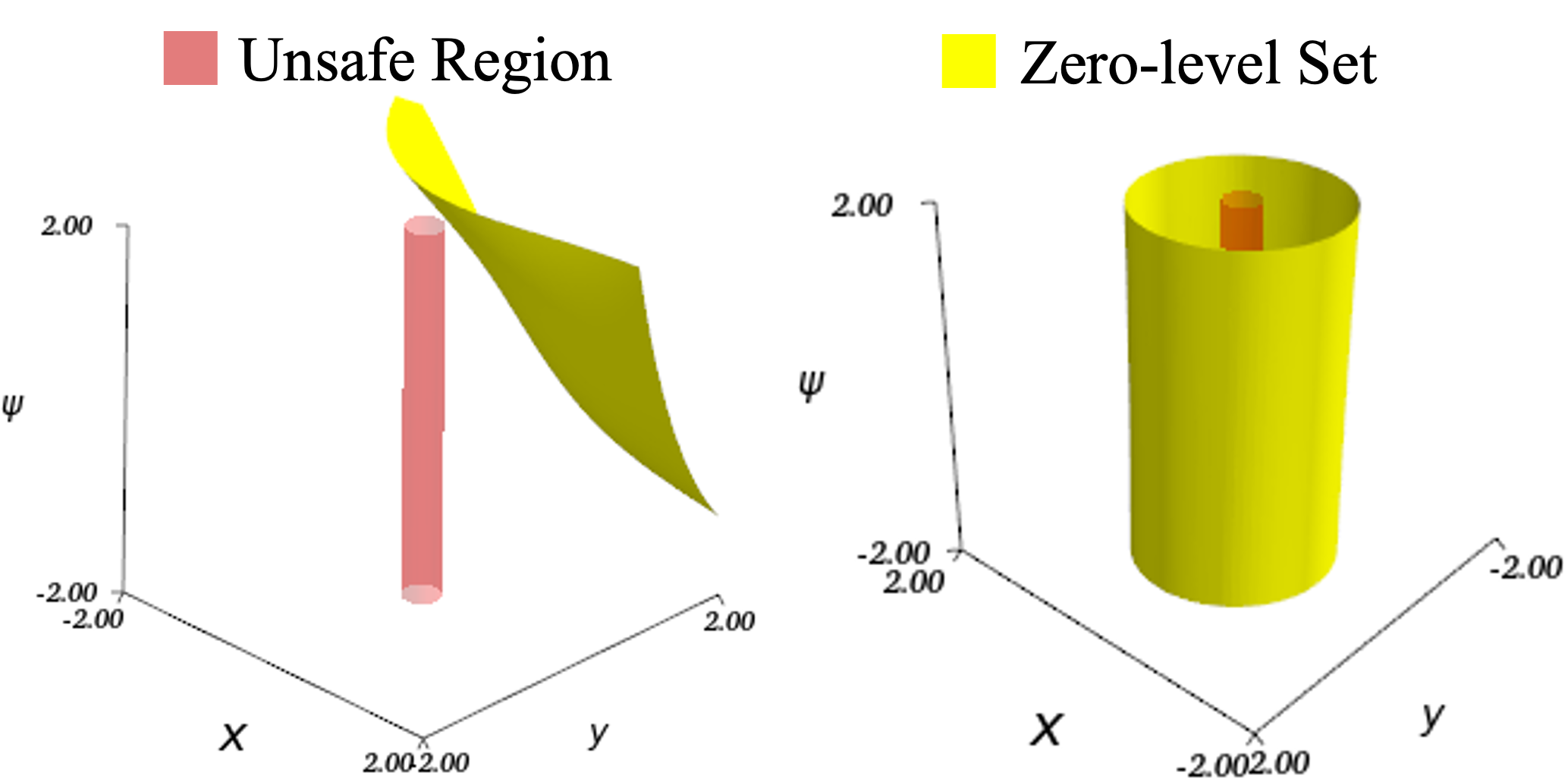}
\caption{The left and right figures show the unsafe region and the zero-level sets of the SNCBF $\tilde{h}$ trained by baseline and the proposed method, respectively. Both zero-level sets (in yellow) do not overlap with the unsafe region in red color. The SNCBF trained by the baseline ensures a guaranteed safe subset over the state space with the ratio $11.8\%$ only, whereas the proposed method ensures a guaranteed safe subset with a ratio up to $77.6\%$. }
    \label{fig:comp_obs}
\end{figure}

We next compare our proposed method with SNCBF trained with baseline proposed in \cite{zhang2024fault} assuming given function $h(x)$. The synthesized SNCBF are visualized in \ref{fig:comp_obs}. 
We observe that the synthesized SNCBF $\tilde{h}$ successfully separates the safe and the unsafe region, since both zero-level sets (in yellow) do not overlap with the unsafe region in red color. We also found that the proposed method ensures a larger subset compared with the baseline. The baseline ensures that $11.8\%$ of the state space is safe, while the proposed method ensures that $77.6\%$ of the state space is safe.


\section{Conclusions}
\label{section: Conclusions}
In this paper, we proposed an algorithmic approach to learn a valid continuous-time neural CBF with formal guarantees in a stochastic environment. We constructed a sample-based training framework to train SNCBF and proved that the efficacy of learned SNCBF by enforcing Lipschitz bounds on the neural network, its Jacobian and (trace of) Hessian terms. We further derived the sufficient condition for the safety of SNCBF-based system. The effectiveness of our proposed approach was demonstrated using the simulation study on inverted pendulum and obstacle avoidance. As a part of future work, we plan to extend this framework to account for unknown dynamics and control bounds. We also plan to improve the algorithm in order to iterate on the improving the size of the safe region and making the learnt CBF less conservative. We also plan to perform hardware experiments on robotic systems.

\label{section: References}
\bibliographystyle{IEEEtran}
\bibliography{references.bib}

\newpage
\label{section: Appendix}
\begin{appendices}
\section{Proof of Theorem \ref{thm: effective_weights}}

\begin{proof}
    Consider the dimension of the input ($x$) is $r \times 1$, the dimension of final weight ($\theta_1$) is $1 \times p$, the dimension of pre-final weight ($\theta_{0}$) is $p \times q$ and the dimension of pre-final bias ($b_{0}$) is $p \times 1$. Let us start by analytically differentiating the neural network: 
    \begin{equation*}
    \begin{aligned}
        y &= \theta_1 \phi(\theta_{0} x + b_{0}) \\
        \frac{\partial y}{\partial x} &= \theta_1 \mathsf{diag}(\phi')\theta_{0}\\
        & (\text{Here, } \phi' = \phi'(\theta_{0} x + b_{0}))\\
        \end{aligned}
    \end{equation*}
    The dimension of $\frac{\partial y}{\partial x}$ is $1 \times r$, therefore, its transpose will have the dimension of $r \times 1$. 
    \begin{equation*}
    \begin{aligned}
        (\frac{\partial y_l}{\partial x})^T &= (\theta_1 \mathsf{diag}(\phi')\theta_{0})^T \\
        &= ( (\phi')^T\mathsf{diag}(\theta_1)\theta_{0})^T\\
        (\because \alpha^T \mathsf{diag}(&\beta) = \beta^T \mathsf{diag}(\alpha), \text{if }\alpha, \beta \text{ are same dim vectors}) \\
        &=\underbrace{\theta_{0}^T \mathsf{diag}(\theta_{1})}_{\hat\theta_l} \phi'(\theta_{0} x + b_{0})\\
        \end{aligned}
    \end{equation*}
Now comparing this with the standard neural network we observe that the derivative term ($\frac{\partial y}{\partial x}$) is behaving like a neural network with activation $\hat\phi=  \phi_l'$ and weight parameters $\hat\theta = (\theta_0,\hat\theta_1)$ and $\hat\theta_1$ is defined as:
    \begin{equation*}
        \hat\theta_1 = \theta_{0}^T \mathsf{diag}(\theta_{1})
    \end{equation*}

Therefore, the certificate for  L-Lipschitz continuity of the derivative term ($\frac{\partial y}{\partial x}$) is given by $M_{\hat\phi}(\hat\theta, \Lambda)\succeq0$.

Now, let us analytically calculate $\mathsf{tr}(\sigma^T\frac{\partial^2 y}{\partial x^2}\sigma)$, where $\sigma$ is $r \times r$ diagonal matrix.

\begin{equation*}
    \begin{aligned}
        \frac{\partial y}{\partial x} &= \hat\theta_1 \hat\phi(\theta_{0} x + b_{0}) \\
        \sigma^T\frac{\partial^2 y}{\partial x^2}\sigma &= \sigma^T(\hat\theta_1 \mathsf{diag}(\hat\phi')\theta_{0} )\sigma\\
        \mathsf{tr}(\sigma^T\frac{\partial^2 y}{\partial x^2}\sigma) &= \mathsf{tr}(\sigma^T(\hat\theta_1 \mathsf{diag}(\hat\phi')\theta_{0} )\sigma)\\
    \end{aligned}
\end{equation*}

\begin{equation*}
    =\underbrace{\begin{bmatrix} \sum_{j=0}^{r}\sigma_j^2\theta_1^{j1}\theta_0^{1j} & \dots & \sum_{j=0}^{r}\sigma_j^2\theta_1^{jn}\theta_0^{pj}
    \end{bmatrix}}_{\bar\theta_1} \phi''(\theta_{0} x + b_{0})
\end{equation*}

Again, comparing this with the standard neural network we observe that $\mathsf{tr}(\sigma^T\frac{\partial^2 y}{\partial x^2}\sigma)$ is behaving like a neural network with activation $\bar\phi= \phi''$ and weight parameters $\bar\theta = (\theta_0,\bar\theta_1)$ and $\hat\theta_1$ is defined as:
    \begin{equation*}
        \bar\theta_1 = \begin{bmatrix} \sum_{j=0}^{r}\sigma_j^2\theta_1^{j1}\theta_0^{1j} & \dots & \sum_{j=0}^{r}\sigma_j^2\theta_1^{jn}\theta_0^{pj}
    \end{bmatrix}
    \end{equation*}

Therefore, the certificate for  L-Lipschitz continuity of the $\mathsf{tr}(\sigma^T\frac{\partial^2 y}{\partial x^2}\sigma)$ term is given by $M_{\bar\phi}(\bar\theta, \Lambda)\succeq0$.

\end{proof}

\end{appendices}

\end{document}